\documentclass[sigconf,nonacm]{acmart}
\microtypesetup{expansion=false}
\usepackage{tikz}
\usepackage{pgfplots}
\usepackage{cuted}
\usepackage{float}
\usepackage{algorithm}
\usepackage{algpseudocode}
\usetikzlibrary{arrows.meta,positioning,fit,calc}
\pgfplotsset{compat=1.18}
\makeatletter
\def\fps@figure{t}
\makeatother
\AtBeginDocument{%
  }

\setcopyright{none}

\begin{document}

\title[Progress-JEPA]{Progress-JEPA: Plan-Aware Representation Learning for Latent World Model Predictive Control}

\author{Jiaxin Bai}
\orcid{0000-0002-8985-6467}
\affiliation{%
  \institution{Hong Kong Baptist University}
  \city{Hong Kong}
  \country{China}}
\email{baijiaxin@hkbu.edu.hk}

\author{Jiaxuan Xiong}
\orcid{0009-0009-2396-1967}
\affiliation{%
  \institution{Hong Kong Baptist University}
  \city{Hong Kong}
  \country{China}}
\email{jxxiong@comp.hkbu.edu.hk}

\renewcommand{\shortauthors}{Bai and Xiong}

\begin{abstract}
  Joint-Embedding Predictive Architectures (JEPAs) learn world models by predicting in representation space rather than reconstructing pixels, making them a natural backbone for latent model predictive control from offline demonstration logs.
  JEPA-style training optimizes short-horizon latent prediction, whereas planning requires a multi-step ranking of imagined futures by goal progress.
  Prior JEPA planners often inherit that ranking from embedding geometry, typically latent Euclidean distance, which arises as a byproduct of representation learning rather than as a progress cost mined from the logs.
  We propose \textbf{Progress-JEPA}, which retains the LeWM encoder--predictor backbone and mines a directed temporal cost from reward-free trajectories: same-trajectory step order supplies positive targets, cross-trajectory pairs act as heuristic negatives, and a rollout-consistency term matches the planner horizon.
  The mined supervision serves two roles: as the deployed planning cost when progress is topological, and as a representation signal that improves Euclidean planning when contact geometry dominates.
  Under locked evaluation, deploying the mined cost raises Two-Room success to \(100.0\%\) versus LeWM's \(97.4\%\), while shared Euclidean planning on the same temporally trained checkpoint raises OGB-Cube by \(14.2\) points over LeWM and improves Push-T.
  Against LeWM and the concurrent RC-aux baseline under locked evaluation, Progress-JEPA matches or exceeds both methods on every environment.
  Ablations show that the directed head, cross-trajectory negatives, and rollout consistency each contribute.
  Progress-JEPA narrows the train--plan gap for JEPA world-model planners by discovering temporal progress structure in offline logs and co-designing cost form with plan-time deployment.
  Code is available at \url{https://github.com/HKBU-KnowComp/Progress-JEPA}.
\end{abstract}

\maketitle

\section{Introduction}

World models aim to capture enough structure about an environment that an agent can reason about futures without acting in the real world~\cite{li2026worldmodeltaxonomy}.
In the planner-facing role that motivates this paper, a world model encodes observations, imagines the consequences of candidate action sequences, and ranks those imagined futures by how close they bring the agent to a goal.
When the only available supervision is offline demonstration logs---observation--action trajectories without dense rewards---predictive dynamics alone do not yield a progress signal that the planner can trust.

Joint-Embedding Predictive Architectures (JEPAs) provide a natural foundation for this setting~\cite{lecun2022path,balestriero2025lejepa}.
Rather than reconstructing pixels, a JEPA learns an encoder that maps observations into a latent space and a predictor that forecasts future embeddings from context and actions.
Useful world knowledge resides in the predictive structure of representations: if the model can anticipate what will happen next in embedding space, it has internalized aspects of the world's dynamics without committing to a generative decoder.
LeWM instantiates this recipe for latent control~\cite{maes2026lewm}.
It trains an action-conditioned encoder--predictor with next-latent prediction, regularizes the representation with SIGReg to avoid collapse~\cite{balestriero2025lejepa}, and at test time searches over action sequences with CEM~\cite{deboer2005cem}, scoring each imagined terminal latent by Euclidean distance to a goal embedding.

This design yields a strong predictive backbone, yet it leaves a structural gap between what is learned and what planning needs.
JEPA-style training optimizes short-horizon prediction in latent space; latent model predictive control (MPC) instead needs a multi-step ranking of futures by goal progress.
In LeWM and related JEPA planners, that ranking is typically inherited from embedding geometry: Euclidean distance between latents becomes the de facto planning cost.
Geometry can correlate with progress when local appearance and control advance together, but it is not trained to reflect steps-to-goal, directed reachability, or planner ranking.
Concurrent planning-aware JEPA variants address pieces of this mismatch: value shaping with offline RL~\cite{destrade2026valueguided}, temporal straightening of latent paths~\cite{wang2026temporalstraightening}, horizon-conditioned reachability auxiliaries~\cite{li2026rcaux}, and progression subspaces~\cite{thil2026sdjepa}.
These approaches either introduce value estimators beyond the demonstration logs, reshape geometry without an explicit plan cost, or treat temporal structure as an auxiliary rather than as a discoverable ranking signal for MPC.

\begin{figure}[t]
  \centering
  \includegraphics[width=\columnwidth]{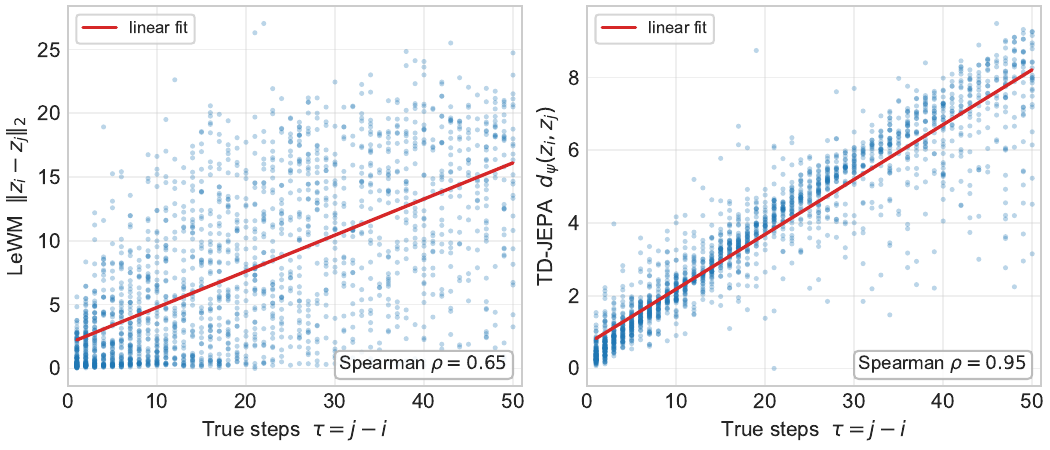}
  \caption{Push-T temporal-gap alignment on held-out demonstration pairs.}
  \Description{Two side-by-side scatter plots with x-axis as true temporal distance and independent per-panel y-axes. Left: LeWM latent L2 (range 0--25) is broadly scattered and weakly monotonic, with a shallow red linear-fit line and Spearman rho 0.65. Right: Progress-JEPA learned d-psi (range 0--8) is tightly increasing along a red linear-fit line with Spearman rho 0.95; no y=x reference is drawn because steps and learned energy do not share units.}
  \label{fig:train_plan_gap}
\end{figure}

We therefore ask: \emph{how can reward-free demonstration trajectories yield temporal progress structure that narrows the train--plan gap for JEPA world-model planners, and when should that structure be deployed as the plan cost versus used to shape representations for geometric planning?}
Figure~\ref{fig:train_plan_gap} makes the ranking side of this gap concrete on Push-T.
On held-out demonstration pairs, LeWM's latent Euclidean distance correlates with temporal separation at Spearman \(\rho=0.65\), whereas a cost mined from the same logs tracks step order at \(\rho=0.95\).
Whether that mined signal also improves control---and under which plan-time cost---is the empirical question addressed below.

\begin{figure*}[t]
  \centering
  \includegraphics[width=0.95\textwidth]{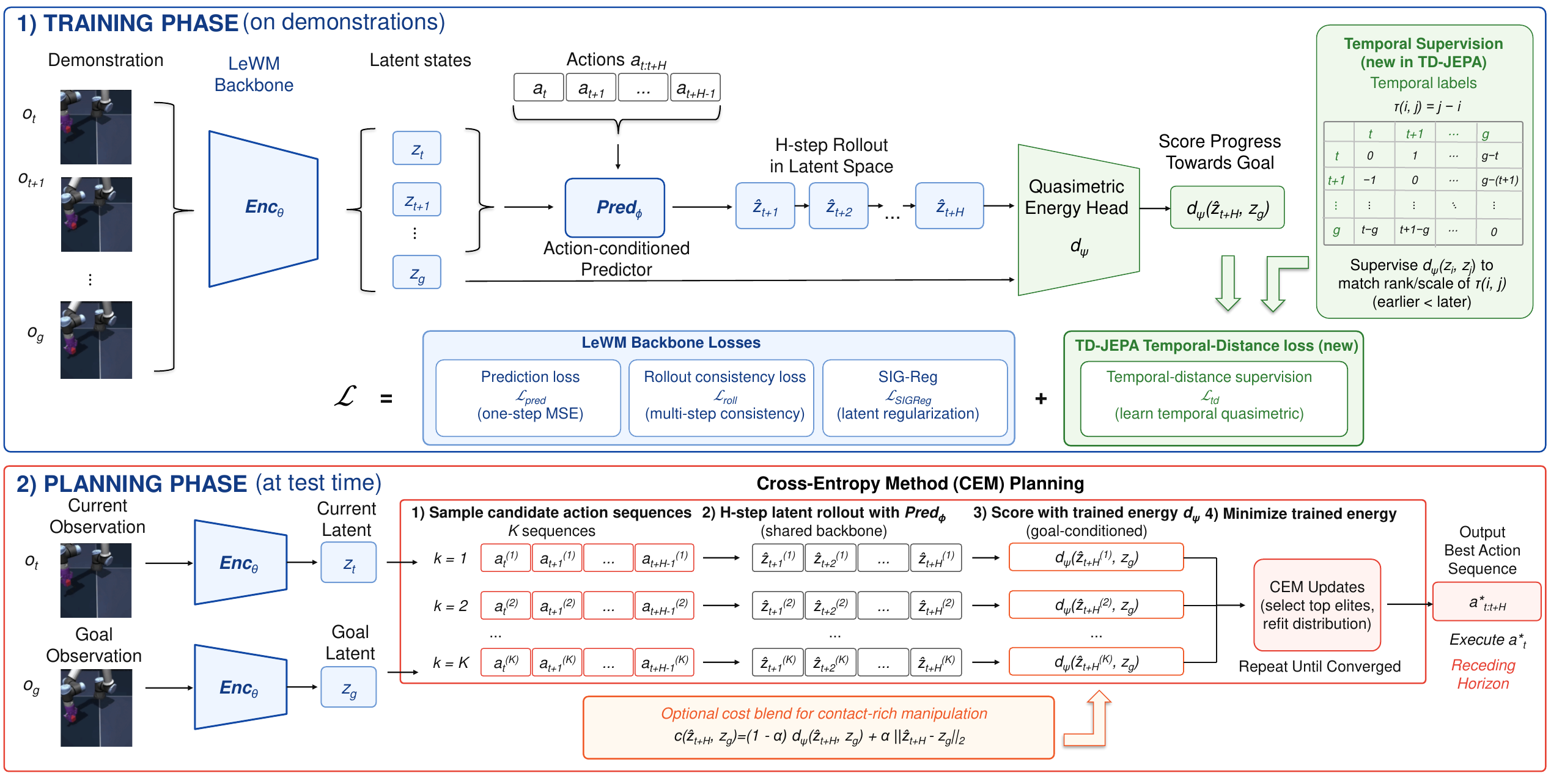}
  \caption{Progress-JEPA training and latent MPC pipeline.
  Observations are encoded into latents; the LeWM predictor rolls out an open-loop trajectory under candidate actions.
  Training keeps the backbone prediction and SIGReg terms, and adds directed temporal-distance supervision (same-trajectory step-order positives and cross-trajectory negatives) plus a horizon-matched loss that supervises recursively predicted latents against encoded future observations.
  At plan time, CEM/iCEM search action sequences that minimize the learned directed cost $d_\psi(\hat z_H,z_g)$ after the same open-loop rollout (latent $\ell_2$ may replace $d_\psi$ on contact-rich tasks).}
  \Description{Pipeline diagram showing observations encoded into latents, a predictor rolling out future latents, a directed cost head scoring the rollout against a goal, and CEM using that cost for planning. Training losses include prediction, rollout consistency, temporal-distance supervision, and SIGReg.}
  \label{fig:pipeline}
\end{figure*}

We propose \textbf{Progress-JEPA} to address this challenge.
Progress-JEPA keeps the LeWM encoder--predictor and SIGReg backbone, and mines a goal-conditioned energy that is low when a rolled-out latent is compatible with reaching the goal~\cite{lecun2006energy}.
Because goal reaching under actions is generally asymmetric, the cost is directed rather than Euclidean~\cite{wang2023quasimetric}: it depends on the ordered pair of current and goal latents, so traveling from~$A$ toward~$B$ need not cost the same as the reverse.
Supervision comes from trajectory mining rather than rewards.
Same-trajectory step order supplies positive temporal targets; cross-trajectory pairs act as heuristic negatives that discourage spuriously low off-trajectory costs; and a rollout-consistency term aligns the learned cost with the open-loop horizon used by the planner (Fig.~\ref{fig:pipeline}).

The mined signal plays two roles, and plan-time cost selection follows task structure.
When progress is mainly topological, as in navigation and reaching, the directed temporal cost is the natural planning objective.
When contact geometry dominates, the same temporally trained checkpoint is planned with latent Euclidean distance, so temporal supervision shapes the representation used by a geometric planner.
We deploy \(d_\psi\) for topology-dominated control and latent \(\ell_2\) for contact-rich control; later sections refer to this design choice rather than re-arguing it.

Locked evaluations support both roles.
Deploying the mined cost improves Two-Room and Reacher over LeWM; shared Euclidean planning on the same temporally trained checkpoint raises OGB-Cube by \(14.2\) points and improves Push-T.
Against LeWM and RC-aux~\cite{li2026rcaux} under locked evaluation, Progress-JEPA matches or exceeds both methods on every environment.
Ablations show that the directed head, cross-trajectory negatives, and rollout consistency each contribute, while contact-phase diagnostics clarify when geometry must be selected at plan time.

\paragraph{Contributions.}
\begin{itemize}
  \item \textbf{A clear diagnosis of the JEPA train--plan gap.} We show that predictive JEPA world models learn dynamics from offline logs, while plan-time scoring often relies on latent embedding geometry rather than an explicitly mined temporal-progress cost, and that a log-derived directed cost closes much of this ranking mismatch (Fig.~\ref{fig:train_plan_gap}).
  \item \textbf{Temporal-distance mining for JEPA planners.} We introduce Progress-JEPA, which keeps the LeWM backbone and adds directed temporal supervision, cross-trajectory negatives, and horizon-matched rollout consistency so that progress structure discovered in demonstrations serves as either a deployed planning cost or a representation signal (Sec.~\ref{sec:temporal_distance}).
  \item \textbf{Evidence that cost form and deployment should be co-designed.} Under matched evaluation, deploying the mined cost helps where topology governs progress, while shared Euclidean planning on the temporally trained checkpoint helps on contact-rich tasks; ablations and contact-phase analyses identify when each choice is appropriate (Tables~\ref{tab:lewm_same_budget}, \ref{tab:nav_cost_deploy}, \ref{tab:ablations}, and~\ref{tab:pusht_contact_gate}).
\end{itemize}

\section{Related Work}

\textbf{Energy-based learning and planning.}
Energy-based models associate low energy with compatible variable assignments and perform inference by energy minimization~\cite{lecun2006energy}.
Latent MPC fits this template by searching for action sequences that minimize terminal energy after imagined rollout.
Value-guided and quasimetric methods learn energies or values from offline data, providing a foundation for goal-conditioned planning costs~\cite{wang2023quasimetric,destrade2026valueguided}.
Progress-JEPA follows the same inference view but changes the knowledge source: instead of a value target, it mines a directed temporal cost from reward-free step-count labels and cross-trajectory negative pairs in demonstration logs~\cite{oord2018cpc,kostrikov2021iql}.

\textbf{JEPA and LeWM.}
Latent-space planning from pixels predates JEPA control.
PlaNet learns stochastic latent dynamics and plans online with predicted rewards, while TD-MPC jointly learns task-oriented dynamics and a terminal value through temporal-difference learning~\cite{hafner2019planet,hansen2022tdmpc}.
JEPAs instead learn predictive representations from data without pixel reconstruction; LeJEPA introduces SIGReg for stable self-supervision, and LeWM combines this recipe with latent planning~\cite{balestriero2025lejepa,maes2026lewm}.
Recent extensions explore relational interventions, online adaptation, and temporal abstraction~\cite{nam2026cjepa,wang2026adajepa,zhang2026hwm,masip2026ffjepa}.
Progress-JEPA leaves the LeWM dynamics backbone unchanged and instead mines the goal-progress cost consumed by the planner.

The central distinction is where task structure is discovered.
PlaNet learns rewards with a reconstructive latent model, and TD-MPC learns reward and value predictions through online interaction.
LeWM removes reward prediction and treats distance in a self-supervised predictive representation as the goal cost.
Progress-JEPA stays reward-free, but makes progress supervision explicit by mining temporal order from demonstration logs.
It therefore targets the missing objective between latent prediction and action search, rather than replacing the dynamics architecture or introducing a learned policy.

\textbf{Value-shaped and temporal visual representations.}
VIP learns a temporally smooth visual embedding whose distance acts as a goal-conditioned reward, and LIV extends value-shaped representation learning to language and image goals~\cite{ma2023vip,ma2023liv}.
TLDR learns temporal-distance-aware representations for goal selection, intrinsic reward, and goal-conditioned policy learning~\cite{bae2024tldr}.
These methods use values or temporal distances to shape rewards and policies.
Progress-JEPA instead mines a directed cost inside an action-conditioned JEPA world model and uses that cost to rank imagined rollouts.
VIP and LIV provide reward signals to a downstream controller; TLDR trains goal-conditioned policies and exploration objectives.
Progress-JEPA stays in the offline, reward-free setting: no reward model and no policy learner, only a temporal cost discovered from logs and consumed by planning.

\textbf{Planning-aware latent geometry.}
Accurate short-horizon prediction is insufficient when MPC searches over multi-step rollouts in a latent space whose metric is not aligned with reachability.
Destrade et al.\ shape JEPA embeddings so a quasimetric distance approximates a goal-conditioned value function learned with IQL, then plan with MPC on that geometry~\cite{destrade2026valueguided}.
Unlike Value-Guided JEPA, Progress-JEPA needs no IQL value estimator: it mines reward-free step-count labels on the LeWM+SIGReg backbone and evaluates task-dependent deployment of \(d_\psi\) versus latent \(\ell_2\) (Table~\ref{tab:novelty_compare}).
Wang et al.\ regularize latent \emph{curvature} so Euclidean distance better tracks geodesic progress after temporal straightening~\cite{wang2026temporalstraightening}.
RC-aux adds multi-horizon open-loop prediction and budget-conditioned reachability supervision on top of LeWM, distinguishing states reachable within the planner's horizon from merely eventually reachable ones~\cite{li2026rcaux}.
SD-JEPA partitions the latent into orthogonal progression and content subspaces, training angular coordinates with a cosine-margin triplet loss while retaining SIGReg on the remainder~\cite{thil2026sdjepa}.
Hierarchical planners and latent-planner JEPAs instead modify inference structure, using coarse latent models or action-free subgoal predictors to reduce long-horizon search difficulty~\cite{zhang2026hwm,masip2026ffjepa}.

\begin{center}
  \captionof{table}{Progress-JEPA vs.\ planning-aware JEPA variants.}
  \label{tab:novelty_compare}
  \setlength{\tabcolsep}{3pt}
  \resizebox{\linewidth}{!}{%
\begin{tabular}{@{}llll@{}}
    \toprule
    Method & Supervision & Directed & Plan use \\
    \midrule
    LeWM~\cite{maes2026lewm} & none & no & latent $\ell_2$ \\
    Value-guided~\cite{destrade2026valueguided} & IQL value & yes & learned value \\
    RC-aux~\cite{li2026rcaux} & horizon budget & no & auxiliary \\
    SD-JEPA~\cite{thil2026sdjepa} & triplet angle & partly & partly \\
    Progress-JEPA & mined step count & yes & $d_\psi$ or $\ell_2$ \\
    \bottomrule
  \end{tabular}%
  }
\end{center}

\textbf{Quasimetric and goal-conditioned RL.}
Quasimetric representations model directed reachability distances; optimal goal-reaching cost-to-go is generally asymmetric~\cite{wang2023quasimetric}.
Value-guided JEPA approximates $-V^\star$ with latent distance via IQL~\cite{destrade2026valueguided,kostrikov2021iql}.
Recent offline GCRL work further strengthens the link between temporal distances and goal reaching: TMD combines contrastive successor-feature learning with quasimetric constraints to recover stitchable temporal distances~\cite{myers2025tmd}, while ProQ uses learned asymmetric distances as directional costs over latent keypoints for long-horizon planning~\cite{kobanda2025proq}.
Temporal representation learning also studies continuous time structure outside JEPA control, including TD InfoNCE for future-event prediction~\cite{zheng2024cdpc} and trajectory encoders that preserve multi-scale spatiotemporal detail in mined mobility logs~\cite{zhou2025blue}.
Progress-JEPA stays in this log-mining view: it regresses a directed temporal cost on demonstration step counts with $i<j$ pairs and does not estimate a reward or value model.

\section{Method}
\label{sec:temporal_distance}

We keep the LeWM encoder--predictor backbone and mine a \textbf{directed temporal cost} $E_\psi \triangleq d_\psi$ from offline demonstration logs.
The method has three parts: the LeWM prediction backbone, a directed cost head calibrated by trajectory step order, and a horizon-matched rollout loss on the same open-loop operator used at plan time.
By default, latent MPC minimizes $d_\psi$ after open-loop rollout; on contact-rich tasks the evaluation protocol instead plans with latent $\ell_2$ on the same temporally trained checkpoint.

\paragraph{Backbone prediction loss.}
Given observation $o_t$ and action $a_t$, the LeWM backbone encodes observations and predicts the next latent:
\begin{equation}
  \hat z_{t+1} = \mathrm{Pred}_{\phi}(\mathrm{Enc}_{\theta}(o_t), a_t),
  \qquad
  z_{t+1} = \mathrm{sg}(\mathrm{Enc}_{\theta}(o_{t+1})).
  \label{eq:jepa_backbone}
\end{equation}
The local dynamics term is the next-latent prediction loss
\begin{equation}
  \mathcal{L}_{\mathrm{pred}} = \|\hat z_{t+1}-z_{t+1}\|_2^2.
  \label{eq:pred_loss}
\end{equation}
Here $\mathrm{sg}$ denotes stop-gradient in our training recipe.
Progress-JEPA retains the LeWM encoder--predictor architecture and SIGReg regularization, then adds planning-specific supervision.

\subsection{Two roles: cost \emph{class} vs.\ cost \emph{calibration}}
\label{sec:two_roles}

Let $z_s$ and $z_g$ denote the current-state and goal latents.
The directed temporal cost $d_\psi(z_s,z_g)$ is a non-negative scalar whose value can depend on the order of its arguments.
We parameterize it with a metric-residual (MRN) head combining a symmetric Euclidean term and an asymmetric residual (Eq.~\eqref{eq:mrn}), and supervise it using forward temporal labels $\tau(i,j)=j-i$ for indices $i<j$ within the same demonstration.
The head specifies the cost's functional form; the temporal labels supervise its values.

\begin{enumerate}
\item \textbf{Directed cost head (geometry class).}
  Goal-reaching under actions is generally \emph{asymmetric}: the cost of reaching $z_g$ from $z_s$ can differ from the reverse cost.
  Optimal goal-reaching value functions and reachability costs are non-negative and asymmetric, motivating quasimetric-style parameterizations~\cite{wang2023quasimetric}.
  Raw latent $\ell_2$ is symmetric ($\|z_s-z_g\|=\|z_g-z_s\|$), so it cannot express this directionality by itself.
  A controlled Push-T ablation keeps the same temporal labels but removes this MRN form: the ablated cost is $\|g_\omega(z_s)-g_\omega(z_g)\|_2$, a symmetric Euclidean distance between learned projection features.
  We evaluate this symmetric temporal-distance head under the same locked Push-T protocol as the other component ablations (Table~\ref{tab:ablations}).

\item \textbf{Temporal-distance supervision (calibration signal).}
  Offline demonstrations supply a reward-free supervision signal: the forward temporal gap is a proxy for steps-to-subgoal \emph{on the data manifold}.
  Eq.~\eqref{eq:td_loss} regresses $d_\psi(z_i,z_j)$ to $\tau(i,j)$, while cross-trajectory hinges penalize spuriously small costs for heuristic negative pairs.
\end{enumerate}

\noindent
\textbf{Composition.}
Temporal labels set the magnitude of $d_\psi$; the quasimetric-style head makes that scalar depend on the order of its arguments.
Together they yield an asymmetric ranking function for goal progress.
Progress-JEPA keeps LeWM's dynamics and anti-collapse terms, and adds a planning term that supervises the \emph{same} $d_\psi$ used in Eq.~\eqref{eq:plan_cost}.

\paragraph{Relation to LeWM planning.}
LeWM uses a symmetric squared latent Euclidean cost, $\|z-z_g\|_2^2$, while Progress-JEPA introduces a directed cost through the MRN head.
Progress-JEPA adds a loss that encourages the plan-time cost to rank same-trajectory pairs by $d_\psi(z_i,z_j)\approx j-i$.
This focus on plan geometry is consistent with recent plan-aware JEPA extensions~\cite{destrade2026valueguided,wang2026temporalstraightening,li2026rcaux}.

\subsection{Energy-based and value-function foundation}
\label{sec:ebm}

Progress-JEPA grounds $d_\psi$ in two established frameworks for plan geometry: energy-based learning~\cite{lecun2006energy} and quasimetric goal-reaching RL~\cite{wang2023quasimetric,destrade2026valueguided}.
This section states the connection and the assumption used by our temporal supervision.

\paragraph{Goal-conditioned energy.}
In the energy-based view~\cite{lecun2006energy}, compatibility of variables $(x,y)$ is scored by an energy $E_\theta(x,y)$ with low values on desirable configurations.
For goal-reaching from latent state $z_s$ to goal $z_g$, define a \emph{directed} energy
\begin{equation}
  E_\psi(z_s, z_g) \triangleq d_\psi(z_s, z_g) \ge 0.
  \label{eq:energy}
\end{equation}
Latent MPC selects actions that minimize terminal energy after rollout:
$a^\star_{1:H} \in \arg\min_{a_{1:H}} E_\psi(\hat z_H(a_{1:H}), z_g)$.
Progress-JEPA \emph{learns} $E_\psi$ and uses the same functional form as the default train-time and plan-time energy.

\paragraph{Link to optimal goal-reaching costs.}
Consider deterministic dynamics with unit cost $\mathbf{1}[s_t \neq g]$ and define $D^\star(s,g)$ as the minimum number of steps needed to reach $g$ from $s$.
On mutually reachable states, this shortest-path cost is non-negative, vanishes on the diagonal, obeys the triangle inequality, and need not be symmetric; it is therefore a quasimetric (or an extended quasimetric if unreachable pairs are assigned $+\infty$)~\cite{wang2023quasimetric}.
Value-guided JEPA methods learn $V_\theta(s,g) \approx V^\star(s,g)$ with implicit Q-learning and set plan cost to $-V_\theta$ or a learned quasimetric surrogate~\cite{destrade2026valueguided}.
Progress-JEPA instead uses explicit temporal labels from offline demonstrations.
For a feasible trajectory segment from $o_i$ to $o_j$, its length $\tau(i,j)=j-i$ upper-bounds the shortest-path distance:
\begin{equation}
  D^\star(o_i,o_j)\leq \tau(i,j)=j-i,
  \qquad
  d_\psi\big(\mathrm{Enc}(o_i),\mathrm{Enc}(o_j)\big)\approx\tau(i,j).
  \label{eq:cost_approx}
\end{equation}
Equality holds only when the demonstrated segment is a shortest path, so we treat demonstration path length as an on-manifold surrogate whose ordering reflects goal progress rather than as $D^\star$ itself.
Eq.~\eqref{eq:cost_approx} is therefore an empirical \emph{ranking-calibration} step.
The MRN head supplies a directed parameterization inspired by the quasimetric structure of $D^\star$~\cite{wang2023quasimetric}; Eq.~\eqref{eq:td_loss} targets $\tau(i,j)=j-i$ in step units as a training signal, and Fig.~\ref{fig:train_plan_gap} shows the resulting monotonic tracking.

\subsection{Directed distance and temporal supervision}
\label{sec:quasimetric}
\label{sec:td_supervision}

We instantiate the directed cost with the metric-residual (MRN) parameterization~\cite{wang2023quasimetric}.
Each latent is mapped to symmetric and asymmetric features $\phi_{\mathrm{sym}}(z)\in\mathbb{R}^{d_s}$ and $\phi_{\mathrm{asym}}(z)\in\mathbb{R}^{d_a}$, and
\begin{equation}
\begin{split}
  d_\psi(z_s, z_g)
  &= \underbrace{\|\phi_{\mathrm{sym}}(z_s)-\phi_{\mathrm{sym}}(z_g)\|_2}_{\text{symmetric component}} \\
  &\quad + \underbrace{\max_{k}\mathrm{ReLU}\!\big(\phi_{\mathrm{asym},k}(z_s)-\phi_{\mathrm{asym},k}(z_g)\big)}_{\text{directed residual}}.
\end{split}
  \label{eq:mrn}
\end{equation}
The head is non-negative by construction and can satisfy $d_\psi(z_s,z_g)\neq d_\psi(z_g,z_s)$.
The symmetric term captures shared configuration factors; the asymmetric residual encodes directed reachability.

Expert demonstrations provide the calibration labels.
For indices $i<j$ on the same trajectory, define $\tau(i,j)=j-i$ as forward steps along the observed segment (not a shortest-path metric).
We minimize
\begin{equation}
\begin{split}
  \mathcal{L}_{\mathrm{td}}
  &= \mathbb{E}_{i<j}\!\left[\rho\!\big(d_\psi(z_i,z_j)-\tau(i,j)\big)\right] \\
  &\quad + \mathbb{E}_{i,\,g'\sim\mathcal{N}}\!\left[\mathrm{ReLU}\!\big(m-\!d_\psi(z_i,z_{g'})\big)\right],
\end{split}
  \label{eq:td_loss}
\end{equation}
where $\rho$ is Smooth L1, $\mathcal{N}$ permutes cross-trajectory goals at the same time index, and $m=\eta(T{-}1)$ is a margin with window length $T$.
The regression term fits on-manifold progress; the hinge discourages spuriously low off-trajectory costs, while admitting false negatives when trajectories share reachable states.
This is regressive EBM calibration without estimating a partition function: positives are shaped toward $\tau(i,j)$, and negatives are pushed above margin $m$~\cite{lecun2006energy,oord2018cpc,destrade2026valueguided}.
We treat demonstration step counts as an on-manifold proxy for progress, not as shortest-path distances.

\paragraph{Train--plan coupling.}
Progress-JEPA trains the energy that serves as the default plan-time cost:
\begin{equation}
  a^\star_{1:H}=\arg\min_{a_{1:H}} d_\psi\!\big(\hat z_H(a_{1:H}), z_g\big),
  \label{eq:plan_cost}
\end{equation}
where $\hat z_H$ is the $H$-step open-loop rollout of the learned predictor.
CEM therefore ranks candidates with the energy shaped by $\mathcal{L}_{\mathrm{td}}$ by default.
For manipulation evaluation, the cost composes this trained energy with latent geometry as $(1{-}\alpha)d_\psi+\alpha\,\ell_2$; Sec.~\ref{sec:pusht_tune} reports this blend as a planning-time diagnostic.

\subsection{Latent MPC with CEM and iCEM}
\label{sec:planner}

At plan time, MPC searches over action sequences that minimize the latent rollout cost in Eq.~\eqref{eq:plan_cost}, executes the first action block, and replans from the resulting observation.
Algorithms~\ref{alg:cem} and~\ref{alg:icem} summarize the two solvers used in our evaluations.
Standard CEM updates a diagonal Gaussian over action sequences from elite moments~\cite{deboer2005cem}.
iCEM adds temporally colored noise, elite reuse across refinements, and momentum-smoothed mean/variance updates~\cite{pinneri2020icem}.
Both call the same latent-rollout cost: open-loop prediction for horizon $H$, then scoring with $d_\psi$ (or latent $\ell_2$ on contact-rich manipulation).

\begin{algorithm}[H]
\caption{Cross-Entropy Method (CEM) for latent MPC}
\label{alg:cem}
\begin{algorithmic}[1]
\Require Context $h_t$, goal latent $z_g$, horizon $H$, candidates $N$, elites $K$, refinements $R$
\State Initialize action mean $\mu^{(0)}$ and standard deviation $\sigma^{(0)}$
\For{$r=0,\ldots,R-1$}
  \For{$n=1,\ldots,N$}
    \State $\epsilon_n \sim \mathcal{N}(0,I)$;\; $A_n \gets \mu^{(r)}+\sigma^{(r)}\odot\epsilon_n$
  \EndFor
  \State Set $A_1\gets\mu^{(r)}$ \Comment{Always evaluate the current mean}
  \State $J_n \gets \Call{LatentRolloutCost}{h_t,z_g,A_n}$ for all $n$
  \State $\mathcal{E}\gets$ indices of the $K$ candidates with lowest $J_n$
  \State $\mu^{(r+1)}\gets \Call{Mean}{\{A_n:n\in\mathcal{E}\}}$;\;
        $\sigma^{(r+1)}\gets \Call{Std}{\{A_n:n\in\mathcal{E}\}}$
\EndFor
\State \Return $\mu^{(R)}$
\end{algorithmic}
\end{algorithm}

\begin{algorithm}[H]
\caption{Improved Cross-Entropy Method (iCEM)}
\label{alg:icem}
\begin{algorithmic}[1]
\Require CEM inputs; color exponent $\beta$, retained elites $M$, smoothing $\gamma$
\State Initialize $\mu^{(0)},\sigma^{(0)}$ and previous elites $\mathcal{E}_{\mathrm{prev}}\gets\varnothing$
\For{$r=0,\ldots,R-1$}
  \State Draw temporally colored noise $\epsilon_{1:N}\sim\mathcal{N}_{\beta}(0,I)$
  \State $A_n\gets\Call{Clip}{\mu^{(r)}+\sigma^{(r)}\odot\epsilon_n}$ for all $n$;\; set $A_1\gets\mu^{(r)}$
  \State Replace up to $M$ candidates with $\mathcal{E}_{\mathrm{prev}}$
  \State $J_n\gets\Call{LatentRolloutCost}{h_t,z_g,A_n}$ for all $n$
  \State $\mathcal{E}\gets$ the $K$ lowest-cost candidates
  \State $\bar{\mu}\gets\Call{Mean}{\mathcal{E}}$;\; $\bar{\sigma}\gets\Call{Std}{\mathcal{E}}$
  \State $\mu^{(r+1)}\gets\gamma\mu^{(r)}+(1-\gamma)\bar{\mu}$;\;
        $\sigma^{(r+1)}\gets\gamma\sigma^{(r)}+(1-\gamma)\bar{\sigma}$
  \State $\mathcal{E}_{\mathrm{prev}}\gets\mathcal{E}$
\EndFor
\State \Return $\mu^{(R)}$
\end{algorithmic}
\end{algorithm}

\subsection{Horizon-aligned rollout consistency}

CEM scores $H$-step open-loop rollouts.
We recursively predict future latents under logged actions and supervise them against the encoded future observations:
\begin{equation}
  \mathcal{L}_{\mathrm{roll}}
  = \frac{1}{H}\sum_{h=1}^{H}\big\|\hat z_{t+h}-\mathrm{sg}(z_{t+h})\big\|_2^2,
  \label{eq:rollout}
\end{equation}
with $H{=}5$ matched to the planner horizon.
Minimizing $\mathcal{L}_{\mathrm{roll}}$ reduces compounding error in the same rollout operator that appears inside Eq.~\eqref{eq:plan_cost}; together with $\mathcal{L}_{\mathrm{td}}$, it couples \emph{temporal} and \emph{spatial} planning structure.

\subsection{Full objective}

The complete Progress-JEPA objective is
\begin{equation}
  \mathcal{L}
  = \mathcal{L}_{\mathrm{pred}}
  + \lambda_{\mathrm{roll}}\mathcal{L}_{\mathrm{roll}}
  + \lambda_{\mathrm{td}}\mathcal{L}_{\mathrm{td}}
  + \lambda_{\mathrm{sig}}\,\mathrm{SIGReg}(z),
  \label{eq:full_loss}
\end{equation}
where $\mathcal{L}_{\mathrm{pred}}$ is the LeWM one-step MSE.
The short-window recipe uses $\lambda_{\mathrm{roll}}{=}0.5$, $\lambda_{\mathrm{td}}{=}1.0$, $\lambda_{\mathrm{sig}}{=}0.09$, prediction window 5, and rollout horizon 5.
SIGReg preserves the anti-collapse prior from LeWM~\cite{balestriero2025lejepa,maes2026lewm}.
At plan time, CEM or iCEM (Algs.~\ref{alg:cem}--\ref{alg:icem}) optimizes Eq.~\eqref{eq:plan_cost} with $d_\psi$ by default; on contact-rich tasks the main protocol uses latent $\ell_2$ on the same checkpoint (Sec.~\ref{sec:pusht_tune}).

\section{Experiments}
\label{sec:experiments}

We address three questions:
(i)~Does temporal-distance mining from reward-free logs improve planning over LeWM and RC-aux~\cite{li2026rcaux} under a fixed evaluation budget?
(ii)~Which mining ingredients drive the gains?
(iii)~When should the mined temporal cost be deployed, and when should latent geometry still be selected?

We study these questions on the four LeWM environments---Two-Room, Reacher, Push-T, and OGBench-Cube---covering navigation, reaching, and contact-rich manipulation.
Progress-JEPA retains the LeWM encoder--predictor backbone and is trained for ten epochs; full hyperparameters appear in Appendix~\ref{app:protocol}.
Unless noted otherwise, primary comparisons report mean$\pm$sample standard deviation (SD) across ten planning seeds, with a fixed checkpoint per method and environment and a shared 50-episode manifest.
Component ablations and phase logs use three planning seeds at fixed checkpoints; rank-correlation diagnostics instead use three independently trained checkpoints.
Additional training repeats in Sec.~\ref{sec:ablations} and Appendix~\ref{app:training_repeats} report SD across independently trained models.
Plan-time defaults follow the two-role design: Progress-JEPA deploys pure $d_\psi$ on Two-Room and Reacher, and latent $\ell_2$ on Push-T and OGB-Cube.
Appendix~\ref{app:training_repeats} additionally evaluates validation-based cost selection on Push-T.
LeWM and RC-aux use latent $\ell_2$ throughout under the same locked manifests.
On Reacher, the planner aggregates terminal and trajectory-mean costs with weight $w{=}0.3$ so that smooth progress is credited; terminal-only scoring is a sensitivity check in Sec.~\ref{sec:reacher_protocol}.

\subsection{Planning Performance}
\label{sec:lewm_same_budget}

Table~\ref{tab:lewm_same_budget} compares Progress-JEPA with LeWM and RC-aux under matched episode manifests and planner budgets.
RC-aux is evaluated with latent $\ell_2$ planning on all four environments, matching its role in Table~\ref{tab:novelty_compare}.

\begin{center}
  \captionof{table}{Locked planning success (\%) over ten planning seeds.
  Progress-JEPA deploys $d_\psi$ on Two-Room/Reacher and shared $\ell_2$ on Push-T/OGB-Cube (Sec.~\ref{sec:experiments}); LeWM and RC-aux use latent $\ell_2$ throughout.}
  \label{tab:lewm_same_budget}
  \setlength{\tabcolsep}{2.5pt}
  \resizebox{\linewidth}{!}{%
  \begin{tabular}{@{}l*{4}{r}@{}}
    \toprule
    Method & Two-Room $\uparrow$ & Reacher $\uparrow$ & Push-T $\uparrow$ & OGB-Cube $\uparrow$ \\
    \midrule
    LeWM & $97.4{\pm}1.3$ & $96.0{\pm}1.9$ & $83.6{\pm}3.2$ & $68.0{\pm}2.8$ \\
    RC-aux~\cite{li2026rcaux} & $98.6{\pm}1.0$ & $96.8{\pm}1.4$ & $81.4{\pm}1.9$ & $81.6{\pm}2.8$ \\
    Progress-JEPA (ours) & $\mathbf{100.0{\pm}0.0}$ & $\mathbf{97.0{\pm}2.4}$ & $\mathbf{86.0{\pm}4.2}$ & $\mathbf{82.2{\pm}2.9}$ \\
    \bottomrule
  \end{tabular}%
  }
\end{center}

Under this locked protocol, Progress-JEPA matches or exceeds both LeWM and RC-aux on every environment.
On Two-Room and Reacher, deploying the mined cost $d_\psi$ yields the gains, showing the value of an explicit temporal plan cost over horizon-budgeted reachability auxiliaries that still plan with geometry.
On Push-T and OGB-Cube, all three methods plan with latent $\ell_2$, so improvements reflect temporal representation learning rather than direct cost deployment.
The largest manipulation gain appears on OGB-Cube, where Progress-JEPA improves over LeWM by $14.2$ percentage points and slightly exceeds RC-aux.
Thus temporal supervision helps in both roles: as a deployed plan cost on topology-dominated tasks, and as a representation signal for geometric planning on contact-rich tasks.

For broader context, Table~\ref{tab:planning_results} places the same locked Progress-JEPA numbers beside published offline RL, imitation learning, and latent world-model baselines under their native protocols.
Those rows are not a matched head-to-head; all LeWM and RC-aux claims in this paper use Table~\ref{tab:lewm_same_budget}.

\begin{center}
  \captionof{table}{Planning success (\%) in literature context. Prior methods are published point estimates under native protocols with different solvers, budgets, and episode draws; Progress-JEPA uses the locked protocol of Sec.~\ref{sec:experiments}. Bold / underline: best / second best per column. Rows are not strictly comparable.}
  \label{tab:planning_results}

  \setlength{\tabcolsep}{3.5pt}
  \resizebox{\linewidth}{!}{%
\begin{tabular}{@{}lllll@{}}
    \toprule
    Method & Two-Room$\uparrow$ & Reacher$\uparrow$ & Push-T$\uparrow$ & OGB$\uparrow$ \\
    \midrule
    GCBC~\cite{ghosh2019gcbc} & $\mathbf{100}$ & --- & $75$ & $\underline{84}$ \\
    IVL~\cite{kostrikov2021iql} & $\mathbf{100}$ & --- & $33$ & $56$ \\
    IQL~\cite{kostrikov2021iql} & $\mathbf{100}$ & --- & $20$ & $64$ \\
    PLDM~\cite{sobal2025pldm} & $\underline{97}$ & $78$ & $78$ & $65$ \\
    DINO-WM~\cite{zhou2025dinowm} & $\mathbf{100}$ & $79$ & $74$ & $\mathbf{86}$ \\
    LeWM~\cite{maes2026lewm} & $87$ & $\underline{86}$ & $\mathbf{96}$ & $74$ \\
    \midrule
    Progress-JEPA (ours) & $\mathbf{100.0{\pm}0.0}$ & $\mathbf{97.0{\pm}2.4}$ & $\underline{86.0{\pm}4.2}$ & $82.2{\pm}2.9$ \\
    \bottomrule
  \end{tabular}%
  }
\end{center}

Relative to published baselines under non-matched protocols, Progress-JEPA achieves the highest reported success on Two-Room and Reacher and remains competitive on OGB-Cube.
We next ablate training ingredients, then ask when pure $d_\psi$ should be deployed.

\subsection{Which Cost Components Matter?}
\label{sec:ablations}

We ablate the directed head, cross-trajectory negatives, and rollout consistency on Push-T (Table~\ref{tab:ablations}).
Each variant uses one checkpoint trained with seed 3072 for ten epochs, continued from its four-epoch checkpoint, and is evaluated with three planning seeds on the same 50 validation episodes.
The columns report a diagnostic blend, pure $d_\psi$, pure latent $\ell_2$, and iCEM; the full-model $\ell_2$ entry of $85.3\%$ is therefore a three-planning-seed diagnostic and need not match the ten-planning-seed $86.0\%$ in Table~\ref{tab:lewm_same_budget}.
The symmetric-head row keeps the temporal regression and hinge but replaces the MRN residual with a Euclidean projection distance, separating cost form from supervision.

\begin{center}
  \captionof{table}{Push-T component ablations under the locked ten-epoch protocol (mean$\pm$sample SD across three planning seeds; one fixed checkpoint per variant).}
  \label{tab:ablations}
  \setlength{\tabcolsep}{2pt}
  \resizebox{\linewidth}{!}{%
  \begin{tabular}{@{}lcccc@{}}
    \toprule
    Variant & Blend & $d_\psi$ & $\ell_2$ & iCEM \\
    \midrule
    Progress-JEPA (full) & $\mathbf{82.7{\pm}6.4}$ & $\mathbf{69.3{\pm}3.1}$ & $\mathbf{85.3{\pm}4.2}$ & $\mathbf{65.3{\pm}6.1}$ \\
    \midrule
    Sym.\ Euclidean TD head & $56.7{\pm}4.2$ & $55.3{\pm}5.0$ & $57.3{\pm}4.2$ & $43.3{\pm}1.2$ \\
    w/o cross-traj.\ hinge & $72.0{\pm}3.5$ & $68.7{\pm}3.1$ & $77.3{\pm}4.2$ & $49.3{\pm}1.2$ \\
    w/o rollout & $62.0{\pm}3.5$ & $56.7{\pm}3.1$ & $60.0{\pm}3.5$ & $44.0{\pm}2.0$ \\
    \bottomrule
  \end{tabular}%
  }
\end{center}

At these fixed checkpoints, each ablation lowers success in every planner setting relative to the full model.
Removing the directed residual produces the largest drop, and removing rollout consistency lowers success more than removing the hinge.
This component comparison supports combining directed temporal supervision with cross-trajectory negatives and horizon-matched rollout consistency.

\paragraph{Separating temporal supervision from rollout consistency.}
We additionally perform a $2\times2$ loss ablation with three independent Push-T training runs per condition (Table~\ref{tab:td_rollout_factorial}).
All arms share the architecture, data split, initialization per training seed, eight-frame windows, and ten-epoch budget, and retain gradients through prediction targets, following LeWM's gradient convention.
We vary $\mathcal{L}_{\mathrm{td}}$ and $\mathcal{L}_{\mathrm{roll}}$ while keeping one-step prediction and SIGReg, and use latent $\ell_2$ planning throughout.
Each model is evaluated with three planning seeds on 200 shared goals; results summarize independent training runs after averaging each model's 600 outcomes.

\begin{center}
  \begin{minipage}{\linewidth}
  \captionof{table}{Push-T loss factorial with prediction-target gradients retained. Success (\%) is mean$\pm$sample SD across three training runs.}
  \label{tab:td_rollout_factorial}
  \setlength{\tabcolsep}{5pt}
  \centering
  \begin{tabular}{@{}ccr@{}}
    \toprule
    TD loss & Rollout loss & Success (\%) \\
    \midrule
    Off & Off & $70.4\pm3.7$ \\
    Off & On  & $57.0\pm43.5$ \\
    On  & Off & $74.3\pm2.3$ \\
    On  & On  & $\mathbf{79.8\pm2.9}$ \\
    \bottomrule
  \end{tabular}
  \end{minipage}
\end{center}

The combined objective achieves the highest mean and lower observed variability than rollout-only across these three training runs.
Appendix~\ref{app:training_repeats} reports all per-run scores, paired intervals, and a separate five-run comparison with RC-aux.

\subsection{When Should the Mined Temporal Cost Be Deployed?}
\label{sec:pusht_tune}

We next fix the Progress-JEPA checkpoint, episodes, solver, and budget, and vary only the plan-time objective among $d_\psi$, latent $\ell_2$, and their blend with $\alpha{=}0.10$.
Figure~\ref{fig:cost_matrix} summarizes the comparison; exact values appear in Appendix Table~\ref{tab:cost_matrix}.
With everything else held fixed, differences in success come from the cost surface seen by CEM.

\begin{figure}[t]
  \centering
  \includegraphics[width=\columnwidth]{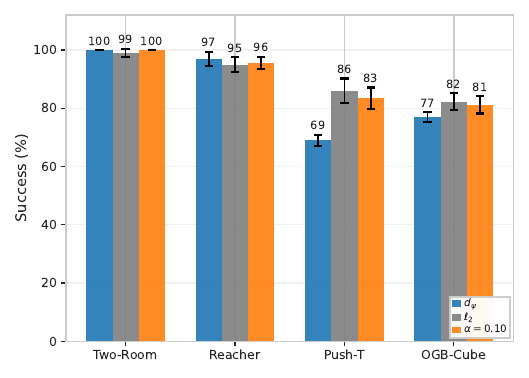}
  \vspace{-0.5cm}
  \caption{Plan-cost comparison with checkpoint, episodes, solver, and budget held fixed (mean$\pm$sample SD over ten planning seeds).}
  \Description{Grouped bar chart of planning success for directed d-psi, latent l2, and their alpha 0.10 blend on Two-Room, Reacher, Push-T, and OGB-Cube. D-psi scores 100.0, 97.0, 69.0, and 77.0. Latent l2 scores 99.0, 95.0, 86.0, and 82.2. The blend scores 100.0, 95.6, 83.4, and 81.2. Error bars show sample standard deviation across ten planning seeds.}
  \label{fig:cost_matrix}
\end{figure}

Plan-cost comparison reveals two regimes.
In navigation-dominated tasks, pure $d_\psi$ matches or exceeds $\ell_2$: Two-Room reaches full success, and Reacher favors the temporal cost.
In contact-rich manipulation, $\ell_2$ outperforms pure $d_\psi$ by a wide margin on Push-T and by a smaller margin on OGB-Cube; a small geometric blend partially closes but does not eliminate the gap.

Table~\ref{tab:nav_cost_deploy} isolates the navigation regime by comparing LeWM's native $\ell_2$ planner with the same Progress-JEPA checkpoint under $d_\psi$ and under $\ell_2$.
Deploying $d_\psi$ exceeds LeWM in both environments, while planning the Progress-JEPA checkpoint with $\ell_2$ falls slightly below LeWM on Reacher.
The navigation gain is therefore not explained by a better Euclidean landscape alone: where topology matters, the temporal cost itself helps.

We further evaluate navigation costs against a grid-based shortest-action-count reference in Two-Room, independently of the trajectory time labels $j-i$.
Across 128 goals and 2,048 state--goal pairs, mean within-goal Spearman correlation is 0.708 for $d_\psi$ versus 0.282 for latent $\ell_2$ on the same checkpoint.
This supports the learned head's alignment with goal-reaching structure.
Appendix~\ref{app:oracle_reference} reports the full comparison, including other learned costs, a privileged-state Euclidean reference, and a grid-resolution check.

\begin{figure}[t]
  \centering
  \includegraphics[width=\columnwidth]{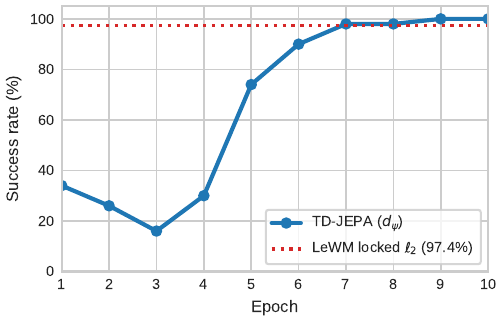}
  \vspace{-0.5cm}
  \caption{Two-Room planning success versus training epoch under pure $d_\psi$ (iCEM-30, seed 3072). The dashed line marks locked LeWM $\ell_2$ success ($97.4\%$).}
  \label{fig:tworoom_curve}
\end{figure}

Figure~\ref{fig:tworoom_curve} shows the Two-Room training trajectory under the deployed $d_\psi$ planner.
Early epochs are unstable under pure temporal scoring, then success climbs from $74\%$ at epoch~5 to full success by epoch~9, surpassing the locked LeWM $\ell_2$ baseline.
This trajectory supports evaluating the final temporally trained checkpoint for navigation deployment under the two-role protocol, rather than selecting an intermediate epoch by peak online score alone.

\begin{center}
  \captionof{table}{Locked navigation costs: LeWM $\ell_2$ versus Progress-JEPA under $d_\psi$ and $\ell_2$ (mean$\pm$sample SD over ten planning seeds).}
  \label{tab:nav_cost_deploy}

  \setlength{\tabcolsep}{3.5pt}
  
\begin{tabular}{@{}llr@{}}
    \toprule
    Environment & Planner setting & Success (\%) \\
    \midrule
    Two-Room & LeWM, $\ell_2$ & $97.4{\pm}1.3$ \\
    Two-Room & Progress-JEPA, $d_\psi$ & $\mathbf{100.0{\pm}0.0}$ \\
    Two-Room & Progress-JEPA, $\ell_2$ & $99.0{\pm}1.4$ \\
    \midrule
    Reacher & LeWM, $\ell_2$ & $96.0{\pm}1.9$ \\
    Reacher & Progress-JEPA, $d_\psi$ & $\mathbf{97.0{\pm}2.4}$ \\
    Reacher & Progress-JEPA, $\ell_2$ & $95.0{\pm}2.5$ \\
    \bottomrule
  \end{tabular}%

\end{center}

\begin{figure*}[t]
  \centering
  \includegraphics[width=\textwidth]{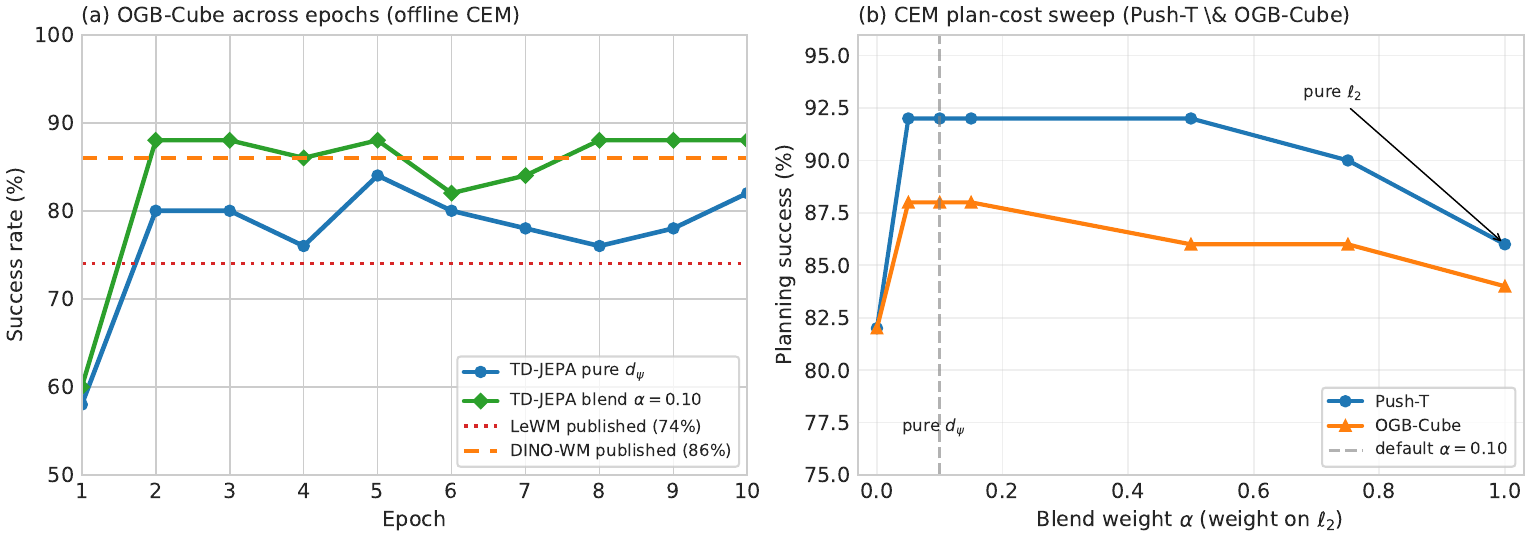}
  \vspace{-0.5cm}
  \caption{Offline CEM diagnostics on manipulation.
  (a)~OGB-Cube success across training epochs under offline-CEM replanning.
  (b)~CEM plan-cost sweep on Push-T and OGB-Cube versus blend weight $\alpha$ in $(1{-}\alpha)\,d_\psi+\alpha\,\ell_2$.}
  \Description{Two-panel figure. Left: OGB-Cube success versus epoch for pure d-psi and alpha 0.10 blend, with LeWM and DINO-WM published baselines. Right: planning success versus blend weight for Push-T and OGB-Cube; both curves rise when a small geometric weight is added.}
  \label{fig:ogb_blend}
  \label{fig:ogb_curve}
  \label{fig:blend_sweep}
\end{figure*}

On manipulation tasks, temporal training improves $\ell_2$ planning even though $d_\psi$ ranks temporal gaps more accurately.
Table~\ref{tab:pusht_rank_split} reports Spearman correlations on Push-T over three independently trained checkpoints with 1{,}000 held-out pairs each: $d_\psi$ exceeds $\ell_2$ on every split, yet $\ell_2$ still plans better in Fig.~\ref{fig:cost_matrix}.
Better temporal ranking is therefore not the same as a better contact-rich plan cost.
The two signals carry different information: $d_\psi$ tracks progress, while $\ell_2$ preserves local geometry useful near contact~\cite{maes2026lewm}.
Offline CEM sweeps in Fig.~\ref{fig:ogb_blend} reinforce the same pattern: manipulation benefits from a geometric plan cost, while temporal training still helps when that geometric cost is used.

\begin{center}
  \captionof{table}{Push-T Spearman correlation with temporal separation (mean$\pm$sample SD across three independent training runs).}
  \label{tab:pusht_rank_split}

  \setlength{\tabcolsep}{4pt}
  \begin{tabular}{@{}lrr@{}}
    \toprule
    Split & $\rho(d_\psi,\tau)$ & $\rho(\ell_2,\tau)$ \\
    \midrule
    All pairs & $\mathbf{0.914 \pm 0.018}$ & $0.785 \pm 0.016$ \\
    Far from endpoint & $\mathbf{0.930 \pm 0.017}$ & $0.789 \pm 0.015$ \\
    Near endpoint/contact & $\mathbf{0.873 \pm 0.037}$ & $0.753 \pm 0.059$ \\
    \bottomrule
  \end{tabular}
\end{center}

The mined temporal signal therefore helps navigation when deployed as $d_\psi$, and helps contact-rich control when used as representation shaping for an $\ell_2$ planner.
The next subsection analyzes \emph{why} pure $d_\psi$ underperforms on Push-T despite stronger temporal ranking.

\subsection{Why Pure Temporal Cost Is Not Enough on Push-T}
\label{sec:pusht_gap}

Unless stated otherwise, Push-T evaluation in this paper uses the primary locked protocol with $\ell_2$ planning in Table~\ref{tab:lewm_same_budget}.
Here we explain why pure $d_\psi$ underperforms that geometric planner: under the locked cost matrix, pure $d_\psi$ reaches $69.0\%$ while latent $\ell_2$ reaches $86.0\%$.
We test whether failures occur before contact (approach) or during post-contact alignment (manipulation).

MPC phase logging on the locked 50-episode manifest answers this (Table~\ref{tab:pusht_phase}).
These logs use three planning seeds at a fixed checkpoint, so per-cost success rates can differ slightly from the ten-planning-seed matrix in Fig.~\ref{fig:cost_matrix}.
Failed $d_\psi$ episodes are almost entirely post-contact, whereas latent $\ell_2$ almost never fails before contact and roughly halves post-contact failures.
Most of the success gap between the two costs arises after contact. On the same checkpoint, latent $\ell_2$ planning achieves fewer post-contact failures, highlighting the importance of cost choice for contact-rich control.

\begin{center}
  \captionof{table}{Where Push-T failures occur under locked CEM (mean$\pm$sample SD of episode counts across three planning seeds at a fixed checkpoint; the same 50 episodes per seed).}
  \label{tab:pusht_phase}

  \setlength{\tabcolsep}{3.5pt}
  \begin{tabular}{@{}lrr@{}}
    \toprule
    Plan cost & Pre-contact fails & Post-contact fails \\
    \midrule
    $d_\psi$ & $1.3{\pm}1.2$ & $14.3{\pm}0.6$ \\
    Latent $\ell_2$ & $0.0{\pm}0.0$ & $6.0{\pm}2.6$ \\
    Blend $\alpha{=}0.10$ & $0.7{\pm}0.6$ & $8.0{\pm}2.6$ \\
    \bottomrule
  \end{tabular}
\end{center}

Start geometry modulates the size of the gap but not its dominant mode (Table~\ref{tab:pusht_gap_geo}).
Across locked ten-planning-seed outcomes, $\ell_2$ uniquely solves far more episodes than $d_\psi$.
The $\ell_2{-}d_\psi$ gap is smaller on above-median close starts and widens on far starts; splits on block--goal pose change the gap much less.
Far starts are harder overall, but Table~\ref{tab:pusht_phase} shows that the dominant failure mode remains post-contact control rather than failing to arrive.

\begin{center}
  \captionof{table}{Push-T success (\%) by start agent--block distance (locked ten-planning-seed protocol).
  The gap column is $\ell_2-d_\psi$.}
  \label{tab:pusht_gap_geo}

  \setlength{\tabcolsep}{3.5pt}
  \begin{tabular}{@{}lrrrr@{}}
    \toprule
    Start stratum & $d_\psi$ & $\ell_2$ & Blend & Gap \\
    \midrule
    All locked episodes & $69.0$ & $86.0$ & $83.4$ & $17.0$ \\
    Agent--block close & $80.8$ & $94.0$ & $90.4$ & $13.2$ \\
    Agent--block far & $57.2$ & $78.0$ & $76.4$ & $20.8$ \\
    \bottomrule
  \end{tabular}
\end{center}

A proximity gate that uses temporal cost far from contact and geometry near contact does not recover most of the $\ell_2$ score without retraining.
Table~\ref{tab:pusht_contact_gate} tests that idea with the same CEM-30 checkpoint and ten planning seeds.
A hard gate improves over pure $d_\psi$ but stays well below pure $\ell_2$; a soft gate matches the $\alpha{=}0.10$ blend and still does not beat geometry alone.
These results favor fixed latent $\ell_2$ scoring over the tested proximity-based cost switches on Push-T.

\begin{center}
  \captionof{table}{Push-T contact-gated planning versus fixed costs (locked CEM-30; mean$\pm$sample SD over ten planning seeds).}
  \label{tab:pusht_contact_gate}
  \setlength{\tabcolsep}{3.5pt}
  \begin{tabular}{@{}lr@{}}
    \toprule
    Plan cost & Success (\%) \\
    \midrule
    Pure $d_\psi$ & $69.0{\pm}1.9$ \\
    Hard contact gate (far $d_\psi$, near $\ell_2$) & $77.6{\pm}5.2$ \\
    Soft contact gate & $84.0{\pm}4.8$ \\
    Fixed blend $\alpha{=}0.10$ & $83.4{\pm}3.7$ \\
    Pure latent $\ell_2$ & $\mathbf{86.0{\pm}4.2}$ \\
    \bottomrule
  \end{tabular}

\end{center}

\subsection{Robustness and Calibration}
\label{sec:reacher_protocol}

How the cost is aggregated over a rollout matters as much as which cost is used.
Table~\ref{tab:reacher_protocol} varies planner aggregation on Reacher under the locked validation manifest.
The primary protocol mixes terminal and trajectory-mean costs with weight $w{=}0.3$, matching temporal-distance supervision; the single-run rates of $96\%$ for Progress-JEPA and $94\%$ for LeWM are consistent with, but not identical to, the ten-seed means in Table~\ref{tab:lewm_same_budget}.
Terminal-only scoring with $w{=}1$ is a sensitivity check that favors geometry and understates the temporal cost.
Under the progressive aggregation used in the primary protocol, CEM and iCEM agree and Progress-JEPA matches or exceeds LeWM.
This choice affects only planner aggregation, not the checkpoint or training objective.

\begin{center}
  \captionof{table}{Reacher solver and rollout-cost aggregation on the locked validation manifest.
  Primary protocol: $w{=}0.3$; $w{=}1$ is a terminal-only sensitivity check.}
  \label{tab:reacher_protocol}
  \small
  \setlength{\tabcolsep}{4pt}
  \resizebox{\linewidth}{!}{%
\begin{tabular}{@{}lrr@{}}
    \toprule
    Protocol & Progress-JEPA & LeWM \\
    \midrule
    CEM-30, $w{=}0.3$ (primary) & $\mathbf{96}$ & $94$ \\
    iCEM-30, $w{=}0.3$ (primary) & $\mathbf{96}$ & $94$ \\
    CEM-30, $w{=}1.0$ (sensitivity) & $74$ & $\mathbf{86}$ \\
    \bottomrule
  \end{tabular}%
  }
\end{center}

Held-out $d_\psi$ rises monotonically with demonstrated step separation (Appendix Fig.~\ref{fig:calibration}), so the head learns the intended ordering even if its absolute scale need not match the true step count.
Together with Table~\ref{tab:pusht_rank_split}, this separates two claims: the cost ranks time well, but good ranking need not suffice for contact-rich control.

\section{Conclusion}

Progress-JEPA narrows the train--plan gap in JEPA world models by mining a directed temporal cost from reward-free demonstration logs on the LeWM encoder--predictor backbone.
The mined signal has two roles: as the deployed plan cost when progress is topological, and as a representation signal when contact-rich control prefers latent geometry.
Locked evaluation supports both choices: deploying $d_\psi$ improves Two-Room and Reacher over LeWM, while shared-$\ell_2$ planning on the same temporally trained checkpoint raises OGB-Cube by $14.2$ points and improves Push-T.
Contact-rich control still prefers geometry at plan time even when $d_\psi$ ranks temporal gaps better, and proximity-gated hybrids do not recover pure geometric scoring.
Component ablations show that the directed residual, cross-trajectory hinge, and rollout consistency each contribute under locked planning.

\begin{acks}
We thank the support of RGC JRFS2526-6S10
\end{acks}

\bibliographystyle{ACM-Reference-Format}
\bibliography{references}

\appendix

\section{Detailed Plan-Cost Results}

Table~\ref{tab:cost_matrix} lists the exact values behind Fig.~\ref{fig:cost_matrix}.
The epoch-10 Progress-JEPA checkpoint, locked 50-episode manifests, environment-specific solvers, and ten planning seeds are held fixed; only the plan cost changes.
The Push-T and OGB-Cube $\ell_2$ cells match the Progress-JEPA row of Table~\ref{tab:lewm_same_budget}.

\begin{center}
  \captionof{table}{Matched plan-cost matrix (mean$\pm$sample SD success over ten planning seeds).}
  \label{tab:cost_matrix}
  \setlength{\tabcolsep}{2.5pt}
  \resizebox{\linewidth}{!}{%
  \begin{tabular}{@{}l*{4}{r}@{}}
    \toprule
    Plan cost & Two-Room $\uparrow$ & Reacher $\uparrow$ & Push-T $\uparrow$ & OGB-Cube $\uparrow$ \\
    \midrule
    Directed temporal $d_\psi$ & $\mathbf{100.0{\pm}0.0}$ & $\mathbf{97.0{\pm}2.4}$ & $69.0{\pm}1.9$ & $77.0{\pm}1.7$ \\
    Latent $\ell_2$ & $99.0{\pm}1.4$ & $95.0{\pm}2.5$ & $\mathbf{86.0{\pm}4.2}$ & $\mathbf{82.2{\pm}2.9}$ \\
    Blend $\alpha{=}0.10$ & $\mathbf{100.0{\pm}0.0}$ & $95.6{\pm}2.1$ & $83.4{\pm}3.7$ & $81.2{\pm}3.0$ \\
    \bottomrule
  \end{tabular}%
  }
\end{center}

\section{Temporal-Cost Calibration}
\label{app:calibration}

\begin{figure}[t]
  \centering
  \includegraphics[width=0.92\columnwidth]{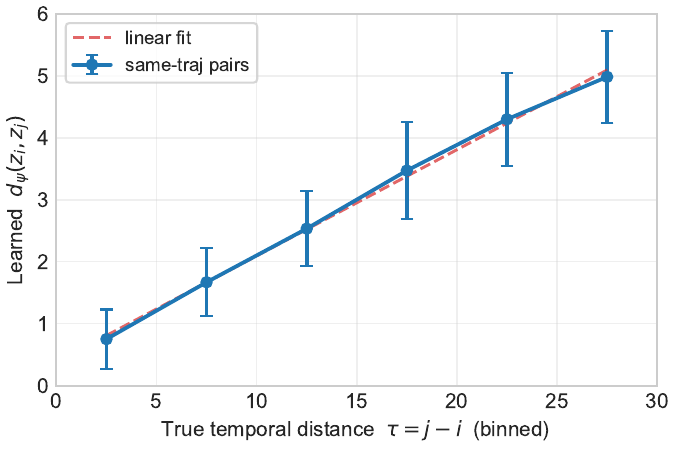}
  \caption{Learned $d_\psi$ versus held-out temporal separation.}
  \Description{Line plot of mean learned d-psi versus binned temporal separation. The mean cost rises monotonically across bins, with error bars showing within-bin variation.}
  \label{fig:calibration}
\end{figure}

On held-out same-trajectory pairs, mean $d_\psi$ rises monotonically with demonstrated step separation.
Absolute scale need not match the true step count; the planner uses the induced ranking.

\section{Training and Evaluation Protocol}
\label{app:protocol}

\paragraph{Training.}
The short-window recipe uses $N_{\mathrm{pred}}{=}5$ (window $T{=}8$ with history size 3), rollout horizon 5, $\lambda_{\mathrm{roll}}{=}0.5$, $\lambda_{\mathrm{td}}{=}1.0$, and $\lambda_{\mathrm{sig}}{=}0.09$.
Directed head: hidden dimension 512; $\phi_{\mathrm{sym}},\phi_{\mathrm{asym}}\in\mathbb{R}^{128}$.
All runs share a ViT-tiny encoder (${\sim}$15M parameters, patch 14, image 224), 192-dimensional latent, history size 3, AdamW with learning rate $5{\times}10^{-5}$ and weight decay $10^{-3}$, bf16 mixed precision, and 10 epochs.
Push-T uses effective batch 128 ($32{\times}4$ accumulation).

\paragraph{Planning defaults.}
MPC uses horizon $H{=}5$, goal offset $G{=}25$, 50 episodes, and 300 candidate action sequences per CEM/iCEM iteration unless noted.
Two-Room: iCEM-30 with pure $d_\psi$ and terminal weight $w{=}1$.
Reacher: iCEM-30 with pure $d_\psi$ and $w{=}0.3$ (terminal mixed with trajectory-mean cost).
Push-T and OGB-Cube locked LeWM/RC-aux comparisons (Table~\ref{tab:lewm_same_budget}): latent $\ell_2$ on the temporally trained checkpoint (CEM-30 and CEM-10; $w{=}1$), matching LeWM's geometric plan cost.
RC-aux uses the same locked manifests and latent $\ell_2$ planning on all four environments (CEM/iCEM settings as LeWM), with multi-horizon rollout and budgeted-reachability training losses as in~\cite{li2026rcaux}.
Table~\ref{tab:planning_results} places the same locked Progress-JEPA numbers in literature context beside published point estimates; it is not a matched head-to-head.
Blend $\alpha{=}0.10$ and pure $d_\psi$ appear in the matched cost matrix and contact diagnostics (Fig.~\ref{fig:cost_matrix}, Tables~\ref{tab:cost_matrix} and~\ref{tab:pusht_contact_gate}).
Push-T LeWM uses the released checkpoint; other environments use our LeWM checkpoints.
iCEM uses \texttt{noise\_beta}${=}2.0$, smoothing coefficient $0.1$, and \texttt{n\_elite\_keep}${=}5$.

\paragraph{Statistical scope.}
For fixed-checkpoint planning-success and phase-count results, $\pm$ denotes the sample SD (denominator $n{-}1$) across planning seeds, with the checkpoint and episode manifest held fixed.
The independent-training studies in Table~\ref{tab:td_rollout_factorial} and Appendix~\ref{app:training_repeats} report sample SD across trained models after averaging each model's planning evaluations.
Push-T ablations (Table~\ref{tab:ablations}) use one epoch-10 checkpoint per variant (training seed 3072) and three planning seeds. Phase logs (Table~\ref{tab:pusht_phase}) use the same three planning seeds. Paired LeWM/RC-aux comparisons, the cost matrix, and contact-gated Push-T results use ten planning seeds (Tables~\ref{tab:lewm_same_budget}, \ref{tab:cost_matrix}, and~\ref{tab:pusht_contact_gate}).
Push-T rank correlations (Table~\ref{tab:pusht_rank_split}) instead report mean$\pm$sample SD across three independently trained checkpoints (training seeds 3072, 3073, and 3074), each evaluated on 1,000 held-out pairs.

\section{Independent Training Repeats on Push-T}
\label{app:training_repeats}

\paragraph{Evaluation and statistical units.}
We supplement the fixed-checkpoint comparisons with two studies that vary training seeds.
Both use final epoch-10 models, three planning seeds, and the same frozen manifest of 200 goal-reaching episodes, compared with 50 episodes in Table~\ref{tab:lewm_same_budget}.
For each model, we average the 600 binary outcomes before computing the mean and sample SD across training runs.
Planning uses latent $\ell_2$, CEM with 300 candidates and 30 iterations, horizon 5, and replanning every five steps.
The studies separately evaluate a target-gradient loss factorial and a stop-gradient comparison with RC-aux.

\paragraph{Matched loss factorial.}
The four arms in Table~\ref{tab:td_rollout_factorial} are trained from scratch with shared initial backbone, projector, predictor, and action-encoder parameters within each training seed; the two TD arms also share head initialization.
All use eight-frame windows (three history frames and five prediction steps), microbatch 32 with gradient accumulation 4, and 113,060 optimizer updates over ten epochs.
AdamW uses learning rate $5\times10^{-5}$ and weight decay $10^{-3}$; the active TD, rollout, and SIGReg weights are 1.0, 0.5, and 0.09, respectively.
This factorial retains gradients through both one-step and rollout prediction targets; it therefore uses the rollout objective in Eq.~\eqref{eq:rollout} with the stop-gradient operator removed.
The base arm uses one-step prediction and SIGReg under this shared training recipe.
Table~\ref{tab:factorial_runs} retains all three training runs, including the low-scoring rollout-only run, whose completed training budget and evaluation records were verified.

\begin{center}
  \begin{minipage}{\linewidth}
  \captionof{table}{Per-training-run success (\%) for the target-gradient factorial. Columns align the shared initializations across arms; each entry averages three planning seeds on 200 goals.}
  \label{tab:factorial_runs}
  \setlength{\tabcolsep}{4pt}
  \centering
  \begin{tabular}{@{}lrrr@{}}
    \toprule
    Objective & Run 1 & Run 2 & Run 3 \\
    \midrule
    Base & 72.83 & 72.33 & 66.17 \\
    Base + rollout & 77.83 & 86.17 & 7.00 \\
    Base + TD & 72.50 & 76.83 & 73.50 \\
    Base + TD + rollout & 82.83 & 77.17 & 79.33 \\
    \bottomrule
  \end{tabular}
  \end{minipage}
\end{center}

\paragraph{Paired comparisons.}
Table~\ref{tab:factorial_intervals} reports approximate 95\% percentile bootstrap intervals from 10,000 resamples of training runs, planning seeds, and goals, sharing the sampled indices across arms.
The two prespecified primary contrasts test adding TD to rollout and adding rollout to the base objective; the other two contrasts are secondary.
The intervals describe uncertainty in the loss increments with three training runs.

\begin{center}
  \begin{minipage}{\linewidth}
  \captionof{table}{Paired success differences in the target-gradient factorial (percentage points), with approximate 95\% bootstrap intervals.}
  \label{tab:factorial_intervals}
  \setlength{\tabcolsep}{3pt}
  \centering
  \begin{tabular}{@{}lrr@{}}
    \toprule
    Contrast & Difference & Interval \\
    \midrule
    Full $-$ rollout-only & $+22.8$ & $[-8.3,71.0]$ \\
    Rollout-only $-$ base & $-13.4$ & $[-57.5,14.6]$ \\
    Full $-$ TD-only & $+5.5$ & $[-1.5,12.7]$ \\
    TD-only $-$ base & $+3.8$ & $[-3.6,11.0]$ \\
    \bottomrule
  \end{tabular}
  \end{minipage}
\end{center}

\paragraph{Five-run comparison with RC-aux.}
A separate study compares Progress-JEPA with our RC-aux reimplementation~\cite{li2026rcaux} across five independent training seeds using the stop-gradient prediction recipe.
Both methods use the same data split, eight-frame windows, optimizer, microbatch and accumulation settings, and ten-epoch update budget described above, with their respective auxiliary losses.
Table~\ref{tab:training_repeat_comparison} reports all five runs.
Progress-JEPA obtains $76.4\pm1.8\%$ and RC-aux $74.9\pm3.4\%$; the paired mean difference is $+1.5$ percentage points with an approximate 95\% interval of $[-3.3,6.3]$ using the same bootstrap procedure.
Pairing uses nominal training seeds, planning seeds, and goals; initial parameter tensors were recorded separately for each method.
These results summarize performance across fresh training runs under the shared 200-goal evaluation protocol.

\begin{center}
  \begin{minipage}{\linewidth}
  \captionof{table}{Independent training repeats under the stop-gradient recipe. Each run averages 600 outcomes; the final row reports mean$\pm$sample SD across five trained models.}
  \label{tab:training_repeat_comparison}
  \setlength{\tabcolsep}{5pt}
  \centering
  \begin{tabular}{@{}lcc@{}}
    \toprule
    Training run & Progress-JEPA & RC-aux \\
    \midrule
    1 & 77.00 & 74.50 \\
    2 & 77.83 & 80.67 \\
    3 & 77.67 & 73.00 \\
    4 & 73.50 & 72.00 \\
    5 & 76.17 & 74.50 \\
    \midrule
    Mean $\pm$ SD (\%) & $76.4\pm1.8$ & $74.9\pm3.4$ \\
    \bottomrule
  \end{tabular}
  \end{minipage}
\end{center}

\paragraph{Validation-based cost selection.}
As a supplementary Push-T study, we select the deployment cost separately for each of the five Progress-JEPA models in Table~\ref{tab:training_repeat_comparison}.
We compare latent $\ell_2$ and $d_\psi$ by mean planning success on 50 validation goals using three planning seeds, then freeze the higher-scoring choice before evaluation on 200 separate test goals.
All five models select $\ell_2$, yielding the $76.4\pm1.8\%$ result in Table~\ref{tab:training_repeat_comparison}, versus $70.9\pm2.7\%$ for $d_\psi$ (mean$\pm$sample SD across trained models).
The selected cost is fixed throughout each model's test episodes.

\section{Independent Shortest-Path Reference}
\label{app:oracle_reference}

We evaluate Two-Room cost rankings against shortest action counts on a simulator-validated state graph, independently of demonstration time labels.
We sample 64 goals per room and, for each goal, eight starting states in the same room and eight in the other room, giving 128 goals and 2,048 directed state--goal pairs.
Sampling is independent of model scores.
Each method uses its original epoch-10 checkpoint; learned costs receive rendered RGB observations, while the coordinate-distance reference uses privileged simulator state.

The primary graph uses a 2.5-pixel grid and legal lattice moves of at most 5 pixels per coordinate, matching the simulator's native speed bound.
Each edge represents one action, and shortest-path search computes the minimum action count to the simulator's success region (distance to the goal below 16 pixels).
The reference is exact within this discrete graph.
All 108,618 primary-graph edges were checked against simulator transitions; independent breadth-first search and shortest-path replay provided additional checks.

For each goal, we compute Spearman correlation across its 16 starting states and then average equally over goals (Table~\ref{tab:oracle_reference}).
Approximate 95\% intervals use 5,000 bootstrap resamples of the 128 goal groups with fixed model checkpoints.
The learned head has higher correlation than the latent $\ell_2$ costs in this comparison; the privileged coordinate-distance reference has the highest overall correlation.

\begin{center}
  \begin{minipage}{\linewidth}
  \captionof{table}{Two-Room cost alignment with the grid-based shortest-action-count reference. Mean within-goal Spearman correlation and approximate 95\% goal-bootstrap interval.}
  \label{tab:oracle_reference}
  \setlength{\tabcolsep}{3pt}
  \centering
  \begin{tabular}{@{}lcc@{}}
    \toprule
    Cost & Mean $\rho$ & 95\% interval \\
    \midrule
    Progress-JEPA $d_\psi$ & 0.708 & $[0.670,0.743]$ \\
    Progress-JEPA $\ell_2$ & 0.282 & $[0.224,0.337]$ \\
    LeWM $\ell_2$ & 0.373 & $[0.318,0.426]$ \\
    RC-aux $\ell_2$ & 0.616 & $[0.576,0.653]$ \\
    \midrule
    State Euclidean (privileged) & 0.813 & $[0.772,0.849]$ \\
    \bottomrule
  \end{tabular}
  \end{minipage}
\end{center}

Halving the grid spacing to 1.25 pixels leaves 98.44\% of reference distances unchanged; the remaining pairs differ by one action.
All five costs retain the same ordering by mean correlation, with $d_\psi$ changing from 0.7080 to 0.7082.

\end{document}